\documentclass{article}

\usepackage[utf8]{inputenc}
\usepackage[final]{corl_2024} 

\usepackage{graphics} 
\usepackage{epsfig} 
\usepackage{times} 
\usepackage{amsmath} 
\usepackage{amssymb}  

\usepackage{tabularx}
\usepackage{booktabs}
\usepackage{multicol}
\usepackage{multirow}
\usepackage{floatrow}
\usepackage[]{hyperref}
\usepackage{algorithm}
\usepackage{algpseudocode}

\usepackage{wrapfig}
\usepackage{tikz}
\usetikzlibrary{tikzmark}

\usepackage{xcolor}
\definecolor{codepurple}{rgb}{0.58,0,0.82}

\title{Towards Open-World Grasping with \\ Large Vision-Language Models}

\author{
  Georgios Tziafas\\
  Department of Artificial Intelligence\\
  University of Groningen  \\
  the Netherlands\\
  \texttt{g.t.tziafas@rug.nl} \\
  \And
  Hamidreza Kasaei \\
  Department of Artificial Intelligence\\
  University of Groningen  \\
  the Netherlands\\
  \texttt{hamidreza.kasaei@rug.nl} \\
}

\begin{document}
\maketitle


\begin{abstract}
The ability to grasp objects in-the-wild from open-ended language instructions constitutes a fundamental challenge in robotics.
An open-world grasping system should be able to combine high-level contextual with low-level physical-geometric reasoning in order to be applicable in arbitrary scenarios.
Recent works exploit the web-scale knowledge inherent in large language models (LLMs) to plan and reason in robotic context, but rely on external vision and action models to ground such knowledge into the environment and parameterize actuation.
This setup suffers from two major bottlenecks: a) the LLM's reasoning capacity is constrained by the quality of visual grounding, and b) LLMs do not contain low-level spatial understanding of the world, which is essential for grasping in contact-rich scenarios.
In this work we demonstrate that modern vision-language models (VLMs) are capable of tackling such limitations, as they are implicitly grounded and can jointly reason about semantics and geometry. 
We propose \texttt{OWG}, an open-world grasping pipeline that combines VLMs with segmentation and grasp synthesis models to unlock grounded world understanding in three stages: open-ended referring segmentation, grounded grasp planning and grasp ranking via contact reasoning, all of which can be applied zero-shot via suitable visual prompting mechanisms.
We conduct extensive evaluation in cluttered indoor scene datasets to showcase \texttt{OWG}'s robustness in grounding from open-ended language, as well as open-world robotic grasping experiments in both simulation and hardware that demonstrate superior performance compared to previous supervised and zero-shot LLM-based methods. Project material is available at \href{https://gtziafas.github.io/OWG_project/}{\textcolor{codepurple}{https://gtziafas.github.io/OWG\_project/}}.

\end{abstract}



\section{Introduction}
\begin{wrapfigure}[11]{r}{0.475\textwidth}
    \vspace{-14mm}
    \centering
     \includegraphics[width=0.98\linewidth]{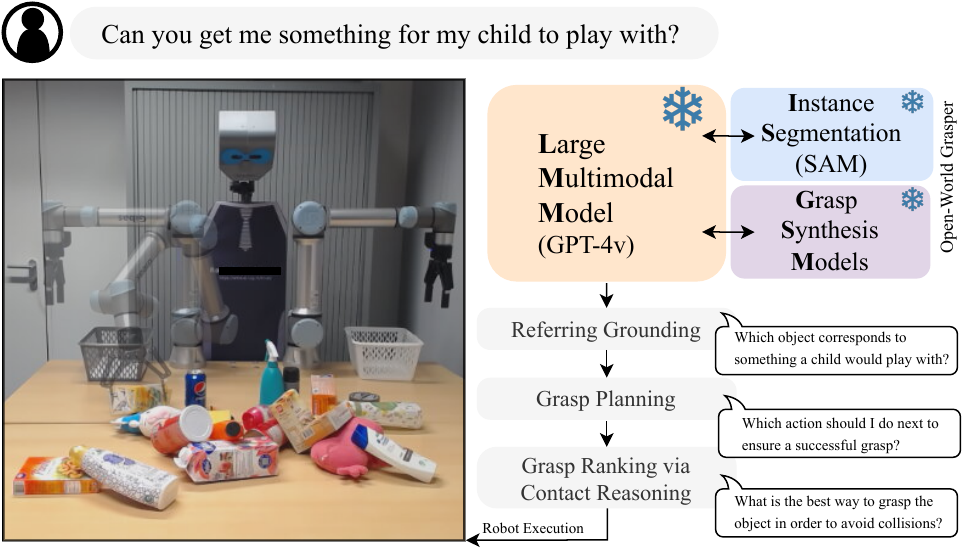}
    \caption{\footnotesize Challenges of open-world grasping tackled with VLMs. The overall pipeline combines VLMs with segmentation and grasp synthesis models to ground open-ended language instructions plan and reason about how to grasp the desired object.}
    \label{fig:fig1}
     \vspace{-2mm}  
\end{wrapfigure}
Following grasping instructions from free-form natural language in open-ended environments is a multi-faceted problem, posing several challenges to robot agents.
Consider the example of Fig.~\ref{fig:fig1}: The robot has to decipher the semantics of the user instruction (i.e., \textit{``what would a child want to play with?"}), recognize the appearing objects and ground the target (i.e., the white toy), reason about the feasibility of the grasp to generate an appropriate plan (i.e., first remove the blocking juice box), and finally select a suitable grasp based on the object geometry and potential collisions.
It becomes clear that to deal with the full scope of open-world grasping, agents should integrate high-level semantic with low-level physical-geometric reasoning, while doing so in a generalizable fashion.

In recent years, Large Language Models (LLMs) \cite{GPT3,GPT4,LLaMAOA,Llama2O,PaLMSL},  have emerged as a new paradigm in robotics and embodied AI, due to their emergent general knowledge, commonsense reasoning and semantic understanding of the world \cite{Gurnee2023LanguageMR,Jiang2019HowCW,Petroni2019LanguageMA,LLMPlannerFG,Ding2022RobotTP}.
This has led to a multitude of LLM-based approaches for zero-shot robotic task planning \cite{DoAI,InnerME,ProgPromptGS,GroundedDG,Text2MotionFN}, navigation \cite{Yu2023L3MVNLL,Zhou2023NavGPTER,Rajvanshi2023SayNavGL,Lin2022ADAPTVN} and manipulation \cite{Stone2023OpenWorldOM,CodeAP,SocraticMC,Instruct2ActMM,ChatGPTFR,VoxPoserC3}, where the LLM decomposes a high-level language instruction into a sequence of steps, therefore tackling complex, long-horizon tasks by composing primitive skills. 
However, a notorious limitation of LLMs is their lack of world grounding — they cannot directly reason about the agent and environment physical state \cite{SetofMarkPU}, and  lack deep knowledge when it comes to low-level, physical properties,
such as object shapes, precise 3D geometry, contact physics and embodiment constraints \cite{LookBB}.
Even when equipped with external visual modules for perceiving the world, the amount of information accessed by the LLM is bottlenecked by the visual model's interface (e.g. open-vocabulary detectors \cite{ViLD,OWL-ViT,MDETRM} cannot reason about object relations such as contacts).
Recently, Large Vision-Language Models (LVLMs) integrate visual understanding and language generation into a unified stream, allowing direct incorporation of perceptual information into the semantic knowledge acquired from language \cite{QwenVLAF, InstructBLIPTG, Liu2023VisualIT,MiniGPT4EV}. 
Preliminary explorations with LVLMs \cite{TheDO} have illustrated two intriguing phenomena, namely: a) by combining LVLMs with segmentation models and constructing suitable visual prompts, LVLMs can unleash extraordinary open-ended visual grounding capabilities \cite{SetofMarkPU}, and b) effective prompting strategies like chain-of-thought \cite{CoT} and in-context examples \cite{GPT3} seem to also emerge in LVLMs.
Motivated by these results, we perform an in-depth study of the potential contributions of LVLMs in open-ended robotic grasping.
In this paper, we propose \textit{Open World Grasper (OWG)}: 
an integrated approach that is applicable zero-shot for grasping in open-ended environments, object catalogs and language instructions.
OWG combines LVLMs with segmentation \cite{SegmentA} and grasp synthesis models \cite{AntipodalRG}, which supplement the LVLM's semantic knowledge with low-level dense spatial inference.
OWG decomposes the task in three stages: open-ended referring segmentation, where the target object is grounded from open-ended language,
(ii) grounded grasp planning, where the agent reasons about the feasibility of grasping the target and proposes a next action, and (iii) grasp ranking, where the LVLM ranks grasp proposals generated from the grasp synthesizer based on potential contacts.

In summary, our contributions are threefold: a) we propose a novel algorithm for grasping from open-ended language using LVLMs, b) we conduct extensive comparisons and ablation studies in real cluttered indoor scenes data \cite{pmlr-v229-tziafas23a, EasyLabelAS}, where we show that our prompting strategies enable LVLMs to ground arbitrary natural language queries, such as open-vocabulary object descriptions, referring expressions and user-affordances, while outperforming previous zero-shot vision-language models by a significant margin, and 
c) we integrate OWG with a robot framework and conduct experiments both in simulation and in the real world, where we illustrate that LVLMs can advance the performance of zero-shot approaches in the open-world setup.

\section{Related Works}
\label{sec:citations}
\textbf{Visual Prompting for Vision-Language Models} Several works investigate how to bypass fine-tuning VLMs, instead relying on overlaying visual/semantic information to the input frame, a practise commonly referred to as \textit{visual prompting}.
Colorful prompting tuning (CPT) is the first work that paints image regions with different colors and uses masked language models to ``fill the blanks" \cite{CPTCP}.
Other methods try to use CLIP \cite{CLIP} by measuring the similarity between a visual prompt and a set of text concepts.
RedCircle \cite{RedCircle} draws a red circle on an image, forcing CLIP to focus on a specific region.
FGVP \cite{FineGrainedVP} further enhances the prompt by specifically segmenting and highlighting target objects.
Recent works explore visual prompting strategies for LVLMs such as GPT-4v, by drawing arrows and pointers \cite{TheDO} or highlighting object regions and overlaying numeric IDs \cite{SetofMarkPU}.
In the same vein, in this work we prompt GPT-4v to reason about visual context while being grounded to specific spatial elements of the image, such as objects, regions and grasps.

\textbf{LLMs/LVLMs in Robotics} Recent efforts use LLMs as an initialization for vision-language-action models \cite{RT1RT,RT2VM}, fine-tuned in robot demonstration data with auxiliary VQA tasks \cite{RT2VM,RT1RT,EmbodiedGPTVP}.
Such end-to-end approaches require prohibitive resources to reproduce, while still struggling to generalize out-of-distribution, due to the lack of large-scale demonstration datasets.
Alternatively, modular approaches invest on the current capabilities of LLMs to decompose language instructions into a sequence of high-level robot skills \cite{zsp,DoAI,InnerME,SocraticMC,GroundedDG}, {or low-level Python programs composing external vision and action models as APIs \cite{ProgPromptGS, Instruct2ActMM,CodeAP,SocraticMC,VoxPoserC3, RobotGPTRM}.
Such approaches mostly focus on the task planning problem, showcasing that the world knowledge built in LLMs enables zero-shot task decomposition, but require external modules \cite{ViLD, OWL-ViT, MDETRM, CLIP} to ground plan steps to the environment and reason about the scene.}
Recent works study the potential of LVLMs for inherently grounded task planning \cite{LookBB, Wake2023GPT4VisionFR, MOKA}. In \cite{Wake2023GPT4VisionFR}, the authors use GPT-4v to map videos of human performing tasks into symbolic plans, but do not consider it for downstream applications.
VILA \cite{LookBB} feeds observation images with text prompts to an LVLM to plan without relying on external detectors.
 However, produced plans are expressed entirely in language and assume an already obtained skill library to execute the plans.
MOKA \cite{MOKA} proposes a keypoint-based visual prompting scheme to parameterize low-level motions, but still relies on external vision models to perform grounding, and does not consider referring expressions and clutter..
 In our work, we use visual marker prompting to leverage LVLMs for the full stack of the open-world grasping pipeline, including grounding referring expressions, task planning and low-level motion parameterization via grasp ranking. 

 \textbf{Semantics-informed Grasping} Most research on grasping assumes golden grounding, i.e., the target object is already segmented from the input scene.
Instead, they focus on proposing 4-DoF grasps from RGB-D views \cite{JacquardAL, EndtoendTD, Cornell, AntipodalRG, gg-cnn, GRConvNetVA, InstancewiseGS}, or 6-DoF poses from 3D data \cite{DexNet2D,GraspNet1BillionAL,acronym2020,Mousavian20196DOFGV,VolumetricGN, Murali20196DOFGF, ContactGraspNetE6}. 
Recently, several works study language-guided grasping in an end-to-end fashion, where a language model encodes the user instruction to provide conditioning for grasping~\cite{An2024LanguagedrivenGD, Lu2023VLGraspA6, pmlr-v229-tziafas23a}. However, related methods typically train language-conditioned graspers that struggle to generalize outside the training distribution.
Another similar line  of works is that of task-oriented grasping~\cite{Ardn2019LearningGA, Murali2020SameOD}, where recent LLM-based methods~\cite{Tang2023GraspGPTLS}
exploit the vast knowledge of LLMs to provide additional semantic context for selecting task-oriented grasps, but do not consider the grounding problem, clutter or referring expressions.
Further, none of the above approaches consider the planning aspect, typically providing open-loop graspers that do not incorporate environment feedback.
In this work, we leverage LVLMs to orchestrate a pipeline for language-guided grasping in clutter, exploiting it's multimodal nature to jointly ground, reason and plan. 

\section{Method}
\label{sec:method}
\subsection{Prerequisites and Problem Statement}
\label{method:problem}

\textbf{Large Vision-Language Models} VLMs receive a set of RGB images of size $H \times W$: $\mathcal{I}_{1:M}, \; \mathcal{I} \in \mathbb{R}^{H \times W \times 3}$ and a sequence of text tokens $\mathcal{T}$, and generate a text sequence $\mathcal{Y}$ of length $L$: $\mathcal{Y}  \doteq w_{1:L}=\left \{  w_1, \dots, w_{L} \right  \}$ from a fixed token vocabulary $w_i \in \mathcal{W}$, such that: $ \mathcal{Y} = \mathcal{F}(\mathcal{I}_{1:M}, \mathcal{T})$.
The images-text pair input $\mathcal{X} = \left\langle \mathcal{I}_{1:M}, \, \mathcal{T} \right\rangle$ is referred to as the \textit{prompt}, with the text component $\mathcal{T}$ typically being a user instruction or question that primes the VLM for a specific task. 

\begin{figure}[t]
    \centering
     \includegraphics[width=0.95\linewidth]{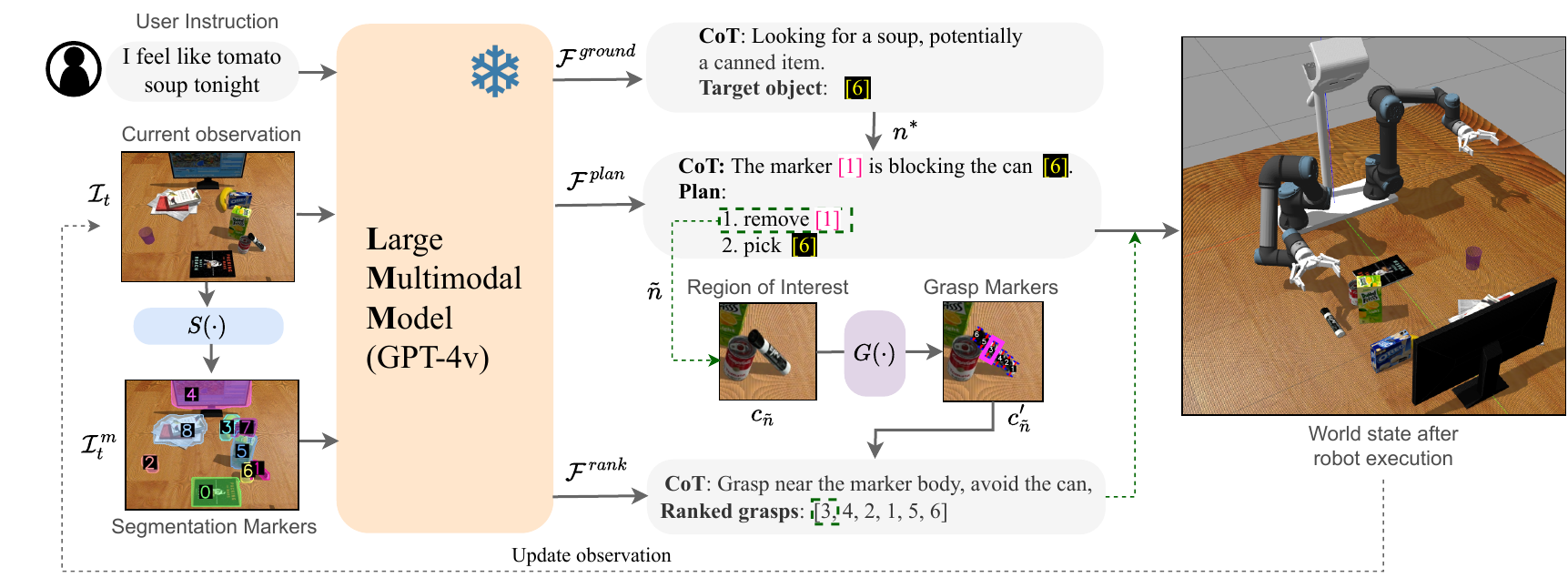}
    \caption{\footnotesize \textbf{Overview of OWG:} Given a user instruction and an observation, OWG first invokes a segmentation model to recover pixel-level masks, and overlays them {with numeric IDs} as visual markers in a new image. Then the VLM subsequently activates three stages: (i) grounding the target object from the language expression in the marked image, (ii) planning on whether it should grasp the target or remove a surrounding object, and (iii) invoking a grasp synthesis model to generate grasps and ranking them according to the object's shape and neighbouring information. The best grasp pose (highlighted here in pink - not part of the prompt) is executed and the observation is updated for a new run, until the target object is grasped. Best viewed in color and zoom.}
    \label{fig:figMain}
\end{figure}

\textbf{Grasp Representations} We represent a grasp via an end-effector gripper pose $\mathcal{G}$, with $\mathcal{G} \in \mathbb{R}^4$ for 4-DoF and $\mathcal{G} \in  \mathbb{R}^6$ for 6-DoF grasping. 
Such representation contains a 3D position and either a yaw rotation or a full SO(3) orientation for 4-DoF and 6-DoF respectively. 
4-DoF grasps assume that the approach vector is calibrated with the camera extrinsics, and hence can be directly drawn as rectangles in the 2D image plane (see bottom of Fig.~\ref{fig:figMain}), which happens to be a favorable representation for VLMs, as grasp candidates can be interpreted as part of the input image prompt.
A motion primitive is invoked to move the arm to the desired gripper pose $\mathcal{G}$, e.g. via inverse-kinematics solvers. \footnote{More sophisticated motion planning algorithms, e.g. with integrated obstacle avoidance, can be utilized orthogonal to our approach.}

\textbf{Problem Statement} Given an RGB-D observation $\mathcal{I}_t \in \mathbb{R}^{H \times W \times 3}$, $\mathcal{D}_t \in \mathbb{R}^{H \times W}$ and an open-ended language query $\mathcal{T}$, which conveys an instruction to grasp a target object, the goal of OWG is to provide a policy $\pi (a_t \mid \mathcal{I}_t, \mathcal{D}_t, \mathcal{T})$.
Assuming $n \in \{1, \dots, N\}$ the $N$ objects that appear in the scene and $n^*$ the target object, then at each time step $t$, the policy outputs a pose for grasping an object: 
$a_t = G_t(n), \; \; G_t(n)=G(n, \mathcal{I}_t, \mathcal{D}_t), \; t=1,\dots,T$, where the last step $T$ always maps to grasping the target object: $a_T=G_T(n^*)$.
We refer to the function $G$ as the \textit{grasp generation} function, which corresponds to a pretrained grasp synthesis network from RGB-D views \cite{AntipodalRG}  \footnote{Other point-cloud \cite{GraspNet1BillionAL} or voxel-based \cite{VolumetricGN} methods for 3D grasp generation can be utilized orthogonal to our approach, which uses single RGB-D view.} 
We note that our policy $\pi$ outputs directly the actual gripper pose $\mathcal{G} = G(n)$, and the object-centric abstraction $n$ is used implicitly (details in next sections).

We wish to highlight that in most grasp synthesis pipelines \cite{AntipodalRG,gg-cnn,EndtoendTD,InstancewiseGS,GRConvNetVA}, it's always $T=1$ and $a_1=G_1(n^*)$, which corresponds to an open-loop policy attempting to grasp the object of interest once.
Our formulation for $T>1$ allows the VLM to close the loop by re-running after each step, which enables visual feedback for planning and recovery from failures / external disturbances.

\subsection{ Pipeline Overview }
\label{method:owg}
OWG combines VLMs with pretrained 2D instance segmentation and grasp synthesis models. 
Segmentation methods like SAM \cite{SegmentA} and its variants \cite{SegmentEE, li2023semantic} have demonstrated impressive zero-shot performane.
Similarly, view-based grasp synthesis networks \cite{GRConvNetVA,gg-cnn,AntipodalRG,EndtoendTD,InstancewiseGS} have also shown to be transferable to unseen content, as they are trained without assumptions of objectness or semantics in their training objectives.
The zero-shot capabilities of these models for low-level dense spatial tasks is complementary to the high-level semantic reasoning capabilities of VLMs, while both use images as the underlying representation, hence offering a very attractive coupling for tackling the open-world grasping problem. 
The overall pipeline can be decomposed in three subsequent stages: (i) {open-ended referring segmentation}, (ii) {grounded grasp planning}, and (iii) {grasp generation and ranking}.
A schematic of OWG is shown in Fig.~\ref{fig:figMain} and described formally in Algorithm~\ref{alg:cap}.
Prompt implementation details can be found in Appendix A.

\begin{figure}[!t]
    \vspace{-12mm}
    \centering
     \includegraphics[width=0.98\linewidth]{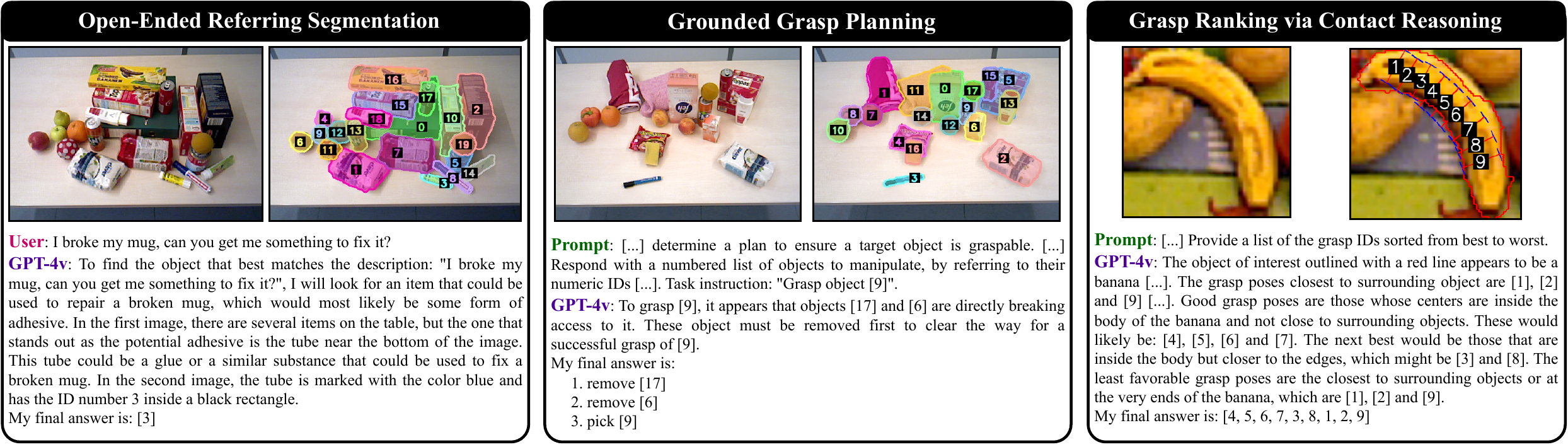}
    \caption{\footnotesize Example GPT-4v responses (from left to right): a) Open-ended referring segmentation, i.e., grounding, b) Grounded grasp planning, and c) Grasp ranking via contact reasoning. We omit parts of the prompt and response for brievity. Full prompts in Appendix A and more example responses in Appendix E.}
    \label{fig:ShowAll}
     \vspace{-2mm}  
\end{figure}

\textbf{Open-ended referring segmentation} In this stage, the target object of interest must be segmented from the input RGB image $\mathcal{I}_t$ given the instruction $\mathcal{T}$.
To enable this, we first run our segmentation model $S:\mathbb{R}^{H \times W \times 3} \rightarrow \{0,1\}^{H \times W}$ and then draw the $N$ generated masks $M_{1:N} = S(\mathcal{I}_t)$ with additional visual markers in a new frame $\mathcal{I}_t^m$.
This step aims to exploit the VLM's OCR capabilities and link each segment in the frame with a unique ID that the VLM can use to refer to it.
After augmenting the image with visual markers, we pass the prompt $<\mathcal{I}_t, \mathcal{I}_t^m, \mathcal{T}>$ to the VLM. 
We refer to this VLM generation as $\mathcal{F}^{ground}$, such that: 
$    
n^* = \mathcal{F}^{ground}(\mathcal{I}_t, \mathcal{I}_t^m, \mathcal{T})$
where $n^*$ the target object and $M_{n^*}$ its segmentation mask.
We note that $\mathcal{T}$ can contain free-form natural language referring to a target object, such as open object descriptions, object relations, affordances etc.

\begin{wrapfigure}[17]{r}{0.6\textwidth}  
    \vspace{-25pt}  
        \begin{algorithm}[H]
        \caption{Open-World Grasper (OWG)}\label{alg:cap}
        \begin{algorithmic}
        \Require Initial observation $(\mathcal{I}_1, \mathcal{D}_1)$, language instruction $\mathcal{T}$, segmentor $S(\cdot)$, grasp generator $G(\cdot)$, VLMs $\mathcal{F}^{ground}, \mathcal{F}^{plan}, \mathcal{F}^{rank}$
        \Ensure $n^* \neq \tilde{n}$
        \State $t \gets 1$
        \While {$n^* \neq \tilde{n}$} 
            \State Generate segmentation masks $M_{1:N}$ with $S(\mathcal{I}_t)$
            \State Draw visual markers from $M_{1:N}$ in a new frame $\mathcal{I}_t^m$
            \State $n^* \gets \mathcal{F}^{ground}(\mathcal{I}_t, \mathcal{I}_t^m, \mathcal{T})$ \Comment{Object of interest}
            \State $\tilde{n} \gets \mathcal{F}^{plan}(\mathcal{I}_t^m, n^*)[0]$ \Comment{Next object to grasp}
            \State $\mathcal{G}_{1:K} \gets G(\mathcal{I}_t,\mathcal{D}_t, M_{\tilde{n}})$ \Comment{Grasp generation}
            \State Crop RoI and draw grasps $c_{\tilde{n}'}$ from $\mathcal{I}_t$, $M_{\tilde{n}}$, $G_{1:K}$
            \State $\mathcal{G'}_{1:K} \gets \mathcal{F}^{rank}(c_{\tilde{n}'})$ \Comment{Grasp ranking}
            \State Execute grasp $\mathcal{G'}_1$
            \State $t \gets t+1$ \Comment{Update observation $\mathcal{I}_t, D_t$}
        \EndWhile
        \end{algorithmic}
        \end{algorithm}
\end{wrapfigure}

\textbf{Grounded grasp planning} This stage attempts to leverage VLM's visual reasoning capabilities in order to produce a plan that maximizes the chances that the target object $n*$ is graspable.
If the target object is blocked by neighboring objects, the agent should remove them first by picking them an placing them in free tabletop space.
Similar to ~\cite{LookBB}, we construct a text prompt that describes these two options (i.e., \textit{remove} neighbor or \textit{pick} target) as primitive actions for the VLM to compose plans from.
We provide the marked image $\mathcal{I}_t^m$ together with the target object $n^*$ (from the previous grounding stage) to determine a plan:
$    p_{1:T} = \mathcal{F}^{plan}(\mathcal{I}_t^m, n^*), \; p_{\tau} \in \{1,\dots,N\} $.
Each $p_{\tau}$ corresponds to the decision to grasp the object with marker ID $n \in \{1,\dots,N\}$. 
As motivated earlier, in order to close the loop, we take the target of the first step of the plan $ \tilde{n} = p_1$ and move to the grasping  stage of our pipeline.

\textbf{Grasp generation and ranking} After determining the current object to grasp $\tilde{n}$, we invoke our grasp synthesis model $G$ to generate grasp proposals.
To that end, we element-wise multiply the mask $M_{\tilde{n}}$ with the RGB-D observation, thus isolating only object $n^*$ in the input frames: $\tilde{\mathcal{I}_t} = \mathcal{I}_t \odot M_{\tilde{n}}, \; \tilde{\mathcal{D}_t} = \mathcal{D}_t \odot M_{\tilde{n}}$.
The grasp synthesis network outputs pixel-level quality, angle and width masks which can be directly transformed to 4-DoF grasps $\mathcal{G}_{1:K} = G(\tilde{\mathcal{I}_t},  \tilde{\mathcal{D}_t})$ \cite{GRConvNetVA,gg-cnn,AntipodalRG}, where $K$ the total number of grasp proposals.
Then, we crop a small region of interest $c_{\tilde{n}}$ around the bounding box of the segment in the frame $\mathcal{I}_t$, from its mask $M_{\tilde{n}}$.
We draw the grasp proposals $\mathcal{G}_{1:K}$ as 2D grasp rectangles within the cropped image $c_{\tilde{n}}$ and annotate each one with a numeric ID marker, similar to the grounding prompt. 
We refer to the marked cropped frame as $c_{\tilde{n}}'$.
Then, we prompt the VLM to rank the drawn grasp proposals:
$
    \mathcal{G}_{1:K}' = \mathcal{F}^{rank}(c_{\tilde{n}}')
$
where the prompt instructs the VLM to rank based on each grasp's potential contacts with neighboring objects.
Finally, the grasp ranked best by the VLM $\mathcal{G}_1'$ is selected and sent to our motion primitive for robot execution.

\section{Experiments}
In this section, we compare the open-ended grounding capabilities of OWG vs. previous zero-shot methods in indoor cluttered scenes (Sec.~\ref{exp1:ocid-vlg}).
Then, we demonstrate its potential for open-world grasping both in simulation and in hardware (Sec.~\ref{exp3:sim}).
Finally, we investigate the effect of several components of our methodology via ablation studies (Sec.~\ref{exp2:ablation}).

\subsection{Open-Ended Grounding in Cluttered Scenes}
\label{exp1:ocid-vlg}
\begin{wraptable}[8]{r}{0.75\textwidth}
     \vspace{-4mm}
    \centering
    \resizebox{\textwidth}{!}{%

    \begin{tabular}{cc|cccccccc}
    \toprule
      \textbf{Method}  &\begin{tabular}{@{}c@{}}\textbf{Found.} \\ \textbf{Model}\end{tabular}  & \textbf{Name} & \textbf{Attribute} & \begin{tabular}{@{}c@{}}\textbf{Spatial} \\ \textbf{Relation}\end{tabular} & \begin{tabular}{@{}c@{}}\textbf{Visual} \\ \textbf{Relation}\end{tabular} & \begin{tabular}{@{}c@{}}\textbf{Semantic} \\ \textbf{Relation}\end{tabular} & \textbf{Affordance} & \begin{tabular}{@{}c@{}}\textbf{Multi-} \\ \textbf{hop}\end{tabular} & \textbf{Avg.} \\
     \midrule 
     ReCLIP~\cite{ReCLIPAS} & CLIP~\cite{CLIP} &   71.4 & 57.7 & 27.3 & 47.4 & 46.2 &62.5 & 20.8 & 47.6\footnotesize{$\pm 17.0$}\\
    RedCircle~\cite{RedCircle} & CLIP~\cite{CLIP} &    52.4 & 53.9 & 18.2 & 42.1 & 46.2 & 18.9 & 12.5 & 34.8\footnotesize{$\pm 16.4$}\\
      FGVP~\cite{FineGrainedVP} & CLIP~\cite{CLIP} &     50.0 & 53.9 & 33.3 & 36.9 & 53.8  & 43.8 & 29.1 & 43.0\footnotesize{$\pm 9.3$}\\
      FGVP$^*$~\cite{FineGrainedVP} & CLIP~\cite{CLIP} &     65.7 & 65.4 & 33.3 & 42.1 & 69.2  & 56.2 & 29.1 & 51.8\footnotesize{$\pm 15.4$}\\
      \midrule 
                  QWEN-VL-2~\cite{QwenVLAF} & QWEN~\cite{QwenVLAF} &     64.3 & 60.9 & 52.4 & 44.0 & 47.1  & 11.9 & 42.1 & 46.1\footnotesize{$\pm 15.9$}\\
                \midrule         
      SoM~\cite{SetofMarkPU}& GPT-4v~\cite{GPT4VisionSC} &   54.8 & 42.3 & 54.6 & 57.9 & 53.9 & 62.5 & 45.8 & 53.1\footnotesize{$\pm 6.4$}\\
      OWG (Ours) & GPT-4v~\cite{GPT4VisionSC} & \textbf{85.7} & \textbf{80.8} & \textbf{75.8} & \textbf{73.7} & \textbf{76.9} & \textbf{93.8} & \textbf{79.2} & \textbf{80.8}\footnotesize{$ \pm 6.4$}\\
     \bottomrule
    \end{tabular}%
  }
    \caption{Zero-shot referring segmentation - mIoU(\%) results per language instruction type for cluttered indoor scenes from OCID \cite{EasyLabelAS}.}
   \label{tab:Ground-SoTa}
    \vspace{-2.2mm}
\end{wraptable}

In order to evaluate the open-ended potential of OWG for grounding, we create a small subset of OCID-VLG test split \cite{pmlr-v229-tziafas23a}, which we manually annotate for a broad range of {grasping instructions}. 
 As we strive for zero-shot usage in open scenes, we mostly experiment with previous visual prompting techniques for large-scale VLMs, such as CLIP \cite{RedCircle,FineGrainedVP,ReCLIPAS}, as well as the recent Set-of-Mark prompting methodology for GPT-4v \cite{SetofMarkPU}, which constitutes the basis of our method.  
We also include comparisons with open-source visually-grounded LVLM QWEN-VL-2~\citep{QwenVLAF}.
Please see Appendix C for details on the test dataset, baseline implementations and more comparative ablations and qualitative results.

{We observe that} both CLIP-based visual prompting techniques and open-source LVLMs are decent in object-based but fail to relate objects from the visual prompts. 
{Even GPT-4v-based SoM prompting method is not directly capable of handling cluttered tabletop scenes from depth cameras, as is evident by the $53.1\%$ averaged mIoU across all query types.
Overall, our OWG-grounder achieves an averaged mIoU score of $80.8\%$, which corresponds to a $27.7\%$ delta from the second best} approach.
Importantly, OWG excels at semantic and affordance-based queries, something which is essential in human-robot interaction applications but is missing from modern vision-language models.
We identify two basic failure modes: a) the LVLM confused the target description with another object, e.g. due to same appearance or semantics, and b) the LVLM reasons correctly about the object and where it is roughly located, but chooses a wrong numeric ID to refer to it.

\subsection{Open-World Grasping Robot Experiments}
\label{exp3:sim}
\begin{wrapfigure}[19]{r}{0.45\textwidth}
    \vspace{-12mm}
    \centering
     \includegraphics[width=\linewidth]{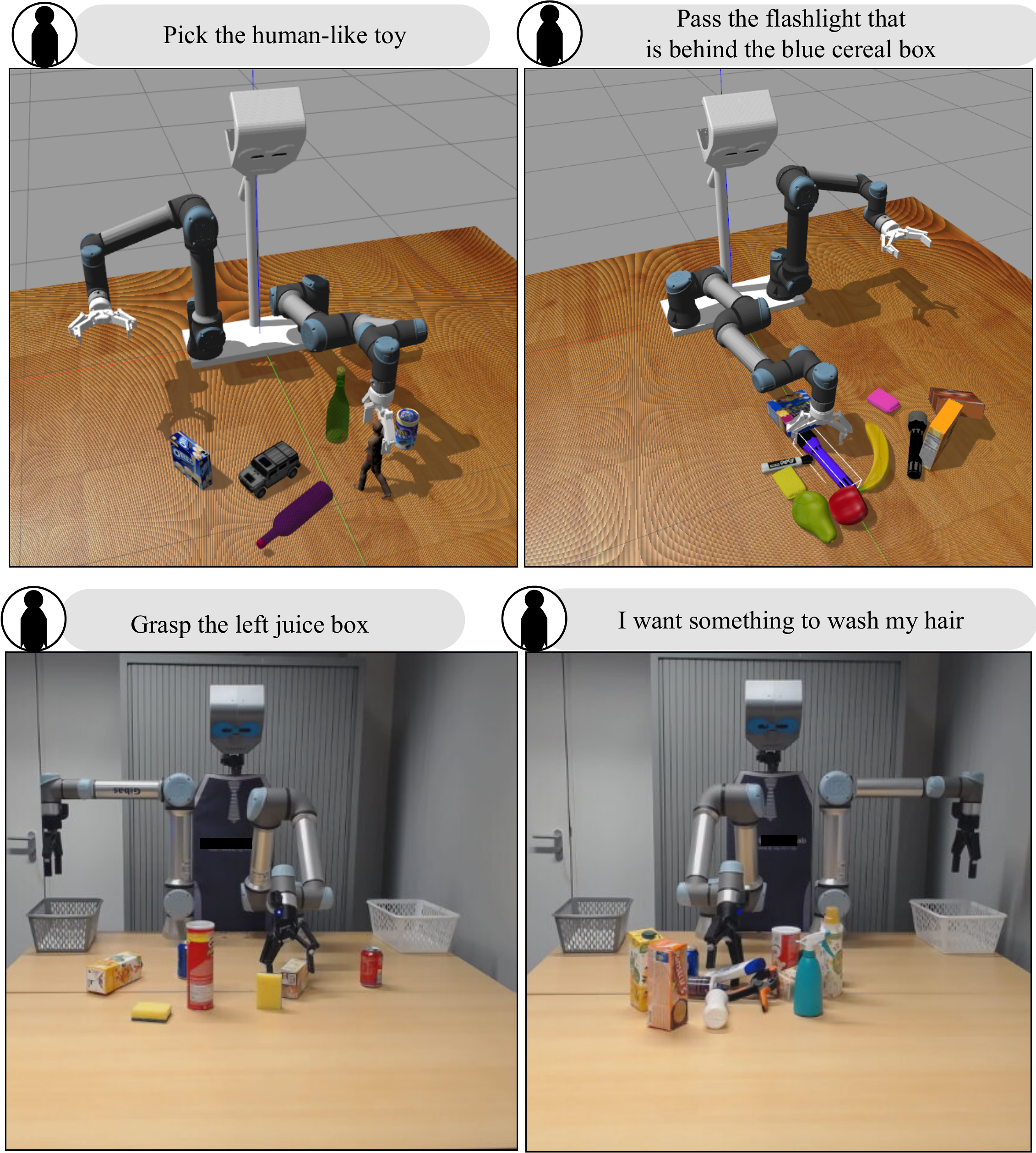}
    \caption{Open-ended language-guided grasping trials in Gazebo \textit{(top)} and real robot \textit{(bottom)}, in isolated \textit{(left column)} and cluttered \textit{(right column)} scenes.}
    \label{fig:rob}
     \vspace{-2.2mm}
\end{wrapfigure}

In this section we wish to evaluate the full stack of OWG, incl. grounding, grasp planning and grasp ranking via contact reasoning, in scenarios that emulate open-world grasping challenges. 
To that end, we conduct experiments in both simulation and in hardware, where in each trial we randomly place 5-15 objects in a tabletop and instruct the robot to grasp an object of interest.
We conduct trials in two scenarios, namely: a) \textbf{isolated}, where all objects are scattered across the tabletop, b) \textbf{cluttered}, where objects are tightly packed together leading to occlusions and rich contacts.
We highlight that object-related query trials contain distractor objects that share the same category with the target object.

\begin{wraptable}[10]{r}{0.5\textwidth}
    \vspace{-1mm} 
    \centering
    \resizebox{\textwidth}{!}{%

    \begin{tabular}{l@{\hskip 0.2in}cc@{\hskip 0.3in}cc@{\hskip 0.3in}cc}
    \toprule
    \textbf{Setup} & \multicolumn{2}{c}{\textbf{CROG}~\cite{pmlr-v229-tziafas23a}} & \multicolumn{2}{c}{\textbf{SayCan-IM}~\cite{InnerME}} & \multicolumn{2}{c}{\textbf{OWG} (Ours)} \\
     \cmidrule(l{2pt}r{14pt}){2-3}
   \cmidrule(l{2pt}r{14pt}){4-5}
   \cmidrule(l{2pt}r{10pt}){6-7}
    & \textit{seen} & \textit{unseen} & \textit{seen} & \textit{unseen} & \textit{seen} & \textit{unseen} \\
    \midrule
    Simulation ($\times 50$) & & & & & & \\
    \quad \textit{-Isolated} & $66.0$ & $36.0$ & $62.0$ & $60.0$ & $78.0$ & $82.0$   \\
    \quad \textit{-Cluttered} & $38.0$ & $22.0$ & $48.0$ & $56.0$ & $62.0$ & $66.0$   \\
     \midrule
    Real-World  ($\times 6$) & & & & & & \\
    \quad \textit{-Isolated} & $50.0$ & $16.6$ & $66.6$ & $33.3$&  $83.3$ & $66.6$  \\
    \quad \textit{-Cluttered} & $16.6$ & $0.0$ & $16.6$ & $16.6$ & $50.0$ & $50.0$  \\
    \bottomrule
    \end{tabular}%
  }
    \caption{\footnotesize Averaged success rates (\%) over simulated and real-world grasping trials. The $\times$ represents number of trials per cell.}
    \label{tab:Ground-Abl}
    \vspace{-1mm}
\end{wraptable}

\textbf{Baselines} We compare with two baselines, namely: a) \textbf{CROG} \cite{pmlr-v229-tziafas23a}, an end-to-end referring grasp synthesis model trained in OCID \cite{EasyLabelAS} scenes, and b) \textbf{SayCan-IM} \cite{InnerME}, an LLM-based zero-shot planning method that actualizes embodied reasoning via chaining external modules for segmentation, grounding and grasp synthesis, while reasoning with LLM chain-of-thoughts \cite{ReActSR}. 
Our choice of baselines aims at showing the advantages of using an LVLM-based method vs. both implicit end-to-end approaches, as well as modular approaches that rely solely on LLMs to reason, with visual processing coming through external tools. See details in baseline implementations in Appendix B.

\textbf{Implementation} Our robot setup consists of two UR5e arms with Robotiq 2F-140 parallel jaw grippers and an ASUS Xtion depth camera.
We conduct 50 trials per scenario in the Gazebo simulator \cite{Gazebo}, using 30 unique object models.
For real robot experiments, we conduct 6 trials per scenario having the initial scenes as similar as possible between baselines. 
In both SayCan-IM and our method, Mask-RCNN \cite{MaskRCNN} is utilized for 2D instance segmentation while GR-ConvNet \cite{AntipodalRG} pretrained in Jacquard \cite{JacquardAL} is used as the grasp synthesis module. 
Our robotic setup is illustrated in Fig.~\ref{fig:rob}, while more details can be found in Appendix B.
To investigate generalization performance, all method are evaluated in {both} scenarios, in two splits: \textit{(i)} \textbf{seen}, where target objects and queries are present in the method's training data or in-context prompts, and \textit{(ii)} \textbf{unseen}, where the instruction refers to objects {that do not appear in CROG's training data or SayCan-IM's in-context prompts}.
 Averaged success rate per scenario is reported, where a trial is considered successful if the robot grasps the object and places it in a pre-defined container position. 

\begin{wrapfigure}[12]{r}{0.5\textwidth}
    \vspace{-1mm}
    \centering
     \includegraphics[width=\textwidth]{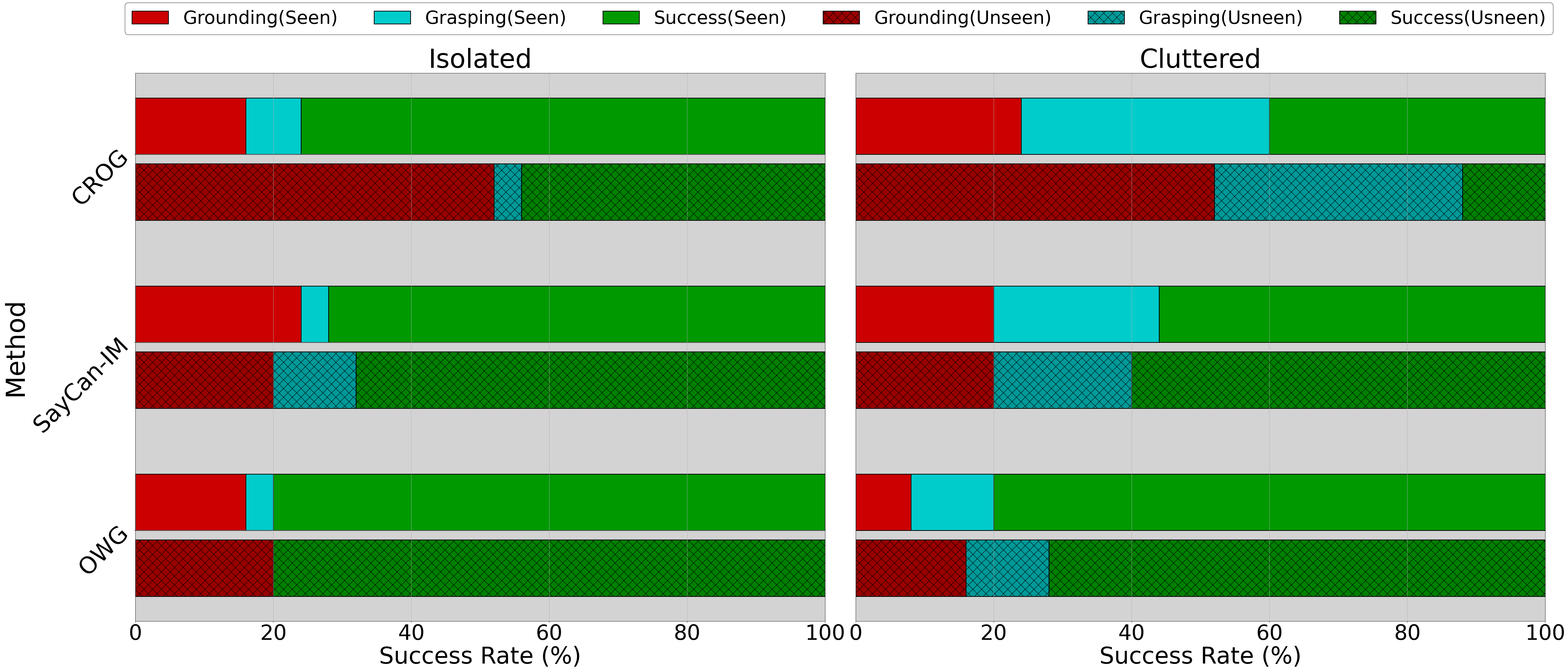}
    \caption{\footnotesize Distribution of failures across grounding and grasping in Gazebo grasping trials for isolated \textit{(left)} and cluttered \textit{(right)}. OWG improves performance across both modes in both setups and test splits.}
    \label{fig:decomp}
     \vspace{-2.2mm}
\end{wrapfigure}

\textbf{Results} We observe that the supervised method CROG struggles when used at unseen data, in both scenarios.
In contrary, both SayCan-IM and OWG demonstrate immunity to seen/unseen objects, illustrating the strong zero-shot capabilities of {LLM-based approaches, which can naturally generalize the concepts of object categories/attributes/relations from language.
SayCan-IM} is {limited} by the external vision models and hence {struggles} in cluttered scenes, {where its} detector sometimes fails to perceive the target object, resulting in {lower final success rates compared to OWG}, especially in the real-world experiments.
OWG consistently outperforms both baselines both in simulation and in the real robot, with an $\sim 15\%$ and $\sim 35\%$ improved averaged success rate respectively.
{In Fig.~\ref{fig:decomp}, we illustrate the decomposition of failures across grounding and grasping in our baselines for $25$ Gazebo trials per scenario, where we automatically test for the target object's grounding results alongside success rate. We observe that OWG consistently reduces the error rates in both grasping and grasping compared to the baselines in all scenarios and test splits.}
We believe that these results are encouraging for the future of LVLMs in robot grasping.

\subsection{Ablation Studies}
\label{exp2:ablation}
In out ablations we wish to answer the following questions: a) What is the bottleneck introduced by the segmentation model in the open-ended grounding performance?, b) What are the contributions of all the different visual prompt elements considered in our work?, and c) What is the contribution of the LVLM-based grasp planning and ranking in robot grasping experiments?
The grounding ablations for the first two questions are organized in Table~\ref{tab:Ground-Abl}, while for the latter in Table~\ref{tab:grasp-Abl}.

\textbf{Instance segmentation bottleneck} We compare the averaged mIoU of our OWG grounder in a subset of our OCID-VLG evaluation data for three different segmentation methods and ground-truth masks.
We employ: a) SAM \cite{SegmentA}, b) the RPN module of the open-vocabulary detector ViLD \cite{ViLD}, and c) the RGB-D two-stage instance segmentation method UOIS \cite{UnseenOI}, where we also provide the depth data as part of the input. 
ViLD-RPN and UOIS both achieve a bit above $70\%$, which is a $\sim 15\%$ delta from ground-truth masks, while SAM offers the best baseline with a $10.8\%$ delta from ground-truth.
Implementation details and related visualizations in Appendix C.

 \begin{wraptable}[12]{r}{0.4\textwidth}
     \vspace{-5mm}
    \centering
    \resizebox{\textwidth}{!}{%

    \begin{tabular}{lc}
    \toprule
    \textbf{Method} & \textbf{mIoU} (\%)  \\
    \midrule
    OWG (w/ \textit{Ground-Truth Mask}) & $86.6$\\
    \quad -w/o reference  & $23.2$ \\
    \quad -w/o number overlay & $54.6$ \\
    \quad -w/o high-res  & $61.3$ \\
    \quad -w/o self-consistency  & $70.9$ \\
    \quad -w/ box  & $74.6$  \\
    \quad -w/o CoT prompt  & $77.6$ \\
    \quad -w/o mask fill  & $81.1$ \\
    \midrule
     SAM~\cite{SegmentA}  & $75.8$ \\
     ViLD-RPN~\cite{OpenvocabularyOD}  & $72.9$ \\
     UOIS~\cite{UnseenOI}  & $71.1$ \\
    \bottomrule
    \end{tabular}}%
    \caption{\footnotesize Grounding ablation studies.}
    \label{tab:Ground-Abl}
    \vspace{-2.2mm}
\end{wraptable}

\textbf{Visual prompt components} Visual prompt design choices have shown to significantly affect the performance of LVLMs. We ablate all components of our grounding prompt and observe the contribution of each one via its averaged mIoU in the same subset as above (see details in Appendix A.2).
The most important prompt component is the reference image, provided alongside the marked image. 
Due to the high clutter of our test scenes, simply highlighting marks and label IDs in a single frame, as in SoM~\cite{SetofMarkPU} hinders the recognition capabilities of the LVLM, with a mIoU drop from $86.6\%$ to $23.2\%$.
Further decluttering the marked image also helps, with overlaying the numeric IDs, using high-resolution images and highlighting the inside of each region mask being decreasingly important. 
Surprisingly, also marking bounding boxes leads to a $12\%$ mIoU drop compared to avoiding them, possibly due to occlusions caused by lots of boxes in cluttered areas.
Finally, self-consistency and chain-of-thought prompting components that were added also improve LVLM's grounding performance by $\sim 16$ and $10\%$ respectively, by ensembling multiple responses and enforcing step-by-step reasoning.

\begin{wraptable}[7]{r}{0.4\textwidth}
     \vspace{-4mm}
    \centering
    \resizebox{\textwidth}{!}{%
    \begin{tabular}{lcc}
    \toprule
    \textbf{Method} & \textbf{Isolated} & \textbf{Cluttered}   \\
    \midrule
    OWG & $84.0$ & $68.0$ \\
    \quad -w/o planning & $80.0$ & $46.0$  \\
    \quad -w/o grasp ranking  & $82.0$ & $60.0$ \\
    \quad -w/o both  & $80.0$ & $42.0$ \\
    \bottomrule
    \end{tabular}}
    \caption{\footnotesize Averaged success rates (\%) over 50 simulated grasping trials per scenario.}
    \label{tab:grasp-Abl}
    \vspace{-1mm}
\end{wraptable}
\textbf{Grasp-Related Ablations} 
We quantify the contribution of our grasp planning and ranking stages in the open-world grasping pipeline, by replicating trials as in the previous section and potentially skipping one or both of these stages.
As we see in Table~\ref{tab:grasp-Abl}, the effect of these components is not so apparent in isolated scenes, as objects are not obstructed by surroundings and hence most proposed grasps are feasible. 
The effect becomes more prominent in the cluttered scenario, where the lack of grasp planning leads to a success rate decrease of $22\%$. 
This is because without grasp planning the agent attempts to grasp the target immediately, which almost always leads to a collision that makes the grasp fail.
Grasp ranking is less essential, as a lot of contact-related information is existent in the grasp quality predictions of our grasp synthesis network. 
However, it still provides an important boost in final success rate ($8\%$ increase).
When skipping both stages, the agent's performance drops drastically in cluttered scenes, as it is unable to recover from grasp failures, and hence always fails when the first attempted grasp was not successful. 

\section{Conclusion, Limitations \& Future Work} 
\label{sec:conclusion}

In this paper we introduce OWG, a novel {system formulation} for tackling open-world grasping.
Our focus is on combining LVLMs with segmentation and grasp synthesis models, and visually prompt the LVLM to ground, plan and reason about the scene and the object grasps.
Our works sets a foundation for enabling robots to ground open-ended language input and close-the-loop for effective {grasp} planning and contact reasoning, leading to significant improvements over previous zero-shot approaches, as demonstrated by empirical evaluations, ablation studies and robot experiments.

\textbf{Limitations}
First, as OWG is a modular approach, it suffers from error cascading effects introduced by the segmentor and grasp synthesis models.
However, improvements in these areas mean direct improvement to the OWG pipeline.
Second, we currently use 4-DoF grasps to communicate them visually to GPT-4v, which constrains grasping to single view. 
In the future we would like to integrate 6-DoF grasp detectors and explore new prompting schemes to aggregate and rank grasp information visually.
Third, our results suggest that LVLMs still struggle to ground complex object relationships.
More sophisticated prompting schemes beyond marker overlaying, {or instruct-tuning in grasp-related data}, might be a future direction for dealing with this limitation.

\clearpage


\bibliography{main}  

\newpage 

\end{document}